\documentclass{article}

\usepackage{microtype}
\usepackage{graphicx}
\usepackage{subfigure}
\usepackage{booktabs} % for professional tables

\usepackage{hyperref}

\usepackage[accepted]{icml2024}

\usepackage{amsmath}
\usepackage{amssymb}
\usepackage{mathtools}
\usepackage{amsthm}

\usepackage[capitalize,noabbrev]{cleveref}

\newcommand{\model}{\textsc{STENCIL}}

\theoremstyle{plain}

\theoremstyle{definition}

\theoremstyle{remark}

\usepackage{subdef}
\usepackage{algorithm}
\usepackage{algorithmic}

\usepackage[textsize=tiny]{todonotes}

\icmltitlerunning{\model: Submodular Mutual Information Based Weak Supervision for Cold-Start Active Learning}

\begin{document}

\twocolumn[
\icmltitle{\model: Submodular Mutual Information Based Weak Supervision for Cold-Start Active Learning}



\icmlsetsymbol{equal}{*}

\begin{icmlauthorlist}
\icmlauthor{Nathan Beck}{equal,yyy}
\icmlauthor{Adithya Iyer}{equal,yyy}
\icmlauthor{Rishabh Iyer}{yyy}
\end{icmlauthorlist}

\icmlaffiliation{yyy}{Department of Computer Science, University of Texas at Dallas}

\icmlcorrespondingauthor{Nathan Beck}{nathan.beck@utdallas.edu}

\icmlkeywords{Active Learning, Cold-Start, NLP, Weak Supervision, Submodular Information Measures, Submodularity, Efficient Deep Learning}

\vskip 0.3in
]



\printAffiliationsAndNotice{\icmlEqualContribution} % otherwise use the standard text.

\begin{abstract}
    As supervised fine-tuning of pre-trained models within NLP applications increases in popularity, larger corpora of annotated data are required, especially with increasing parameter counts in large language models. Active learning, which attempts to mine and annotate unlabeled instances to improve model performance maximally fast, is a common choice for reducing the annotation cost; however, most methods typically ignore class imbalance and either assume access to initial annotated data or require multiple rounds of active learning selection before improving rare classes. We present \model, which utilizes a set of text exemplars and the recently proposed submodular mutual information to select a set of weakly labeled rare-class instances that are then strongly labeled by an annotator. We show that \model\ improves overall accuracy by $10\%-18\%$ and rare-class F-1 score by $17\%-40\%$ on multiple text classification datasets over common active learning methods within the class-imbalanced cold-start setting.
\end{abstract}

\section{Introduction}

In recent years, many natural language processing use cases have been driven by the advent of widely available pre-trained models and subsequent fine-tuning~\cite{min2023recent}. Indeed, this paradigm has become commonplace with the prevalence of large language models, seeing heavy use in cases such as instruction fine-tuning~\cite{zhang2023instruction, ouyang2022training}. The success of this paradigm for downstream tasks, however, relies heavily on annotated data. To scale to today's pre-eminent models, large corpora of supervised data are needed, resulting in prohibitively large annotation costs. One of the most common choices for reducing annotation cost is through the paradigm of active learning, which aims to select the best instances from unlabeled data for annotation that improves model performance maximally fast. Indeed, active learning has been used extensively both in classical machine learning and recent deep learning approaches~\cite{ren2021survey, settles2009active, beck2021effective}. 

Despite the successful use of active learning, complications in the downstream learning environment can prohibit the use of many common methods. One complication is the presence of class imbalance, which tends to reduce the efficacy of active learning methods designed for the general case. Another complication is having to deploy such strategies in cold-start scenarios. Indeed, the majority of published active learning methods assume access to an initial annotated set of data. When both complications are combined, active learning methods for dealing with a \emph{class-imbalanced cold-start} environment are needed.

While the cold-start scenario has been considered for active learning, such methods tend to operate over multiple rounds of selection and retraining~\cite{barata2021active, brangbour2022cold, ni2020merging, hacohen2022active, yuan2020cold, jin2022cold,kothawade2022active}, which addresses class imbalance inefficiently. Due to the nature of cold-start scenarios, such gradual build-up is natural as the active learning strategy learns more about the data landscape before capitalizing on selecting rare-class instances. However, such capitalization can be conducted earlier in cases where prior knowledge can be incorporated, which is often the case in many fine-tuning scenarios and can be conducted via \emph{weak supervision}~\cite{ratner2017snorkel, maheshwari2020semi, maheshwari-etal-2022-learning, rauch2022enhancing}. For example, using active learning for spam detection with unlabeled spam-imbalanced data can be accelerated if the sketch of a spam instance is known. Such a sketch is also easier to provide than in other tasks (such as drawing exemplar images for vision tasks). Hence, \emph{can class-imbalanced cold-start active learning more quickly improve rare-class performance if prior knowledge of the data environment can be provided?}

In this work, we present \model -- \textbf{S}ubmodular mutual informa\textbf{T}ion based w\textbf{E}ak supervisio\textbf{N} for \textbf{C}old start act\textbf{I}ve \textbf{L}earning -- that effectively utilizes prior knowledge of the task via a small exemplar set of rare-class instances. \model\ uses this set to guide the active learning selection by maximizing the Submodular Mutual Information (see Section~\ref{sec:prelim})~\cite{iyer2021submodular,iyer2021generalized} between the set of selected instances and the exemplar set. As we show in Section~\ref{sec:experiments}, \model\ provides an immediate improvement of $10\%-18\%$ overall accuracy and $17\%-40\%$ rare-class F-1 score with a single round of selection compared to other methods. \model\ achieves this performance gain using as few as 15-25 exemplar instances, which are easy to provide in a variety of NLP fine-tuning tasks. Hence, \model\ offers a sound and complimentary strategy for the cold-start round of class-imbalanced active learning, especially for NLP fine-tuning tasks where \model's weak supervision can be used.

\iffalse
\subsection{Related Work}

\begin{itemize}
    \item \url{https://arxiv.org/pdf/2107.07724.pdf} Look at AL in 3 stages: cold, warmup, and hot. Apply ODAL in warmup and hot stages. We're different in that we start directly from cold scenario (no labeled data) and guide selection based on simple exemplars. (Maybe add unsup + outlier detection) (Maybe add coreset as well).
    \item \url{https://arxiv.org/pdf/2201.10227.pdf} Do AL via label propagation objective + cluster ranking. A proxy model is used to first get data, whereby a separate classifier is then subsequently trained on the labeled data. Drawbacks: Doesn't immediately produce downstream classifier, still does (almost) random selection for very first oracle query (not guaranteed to give rare-class data, which again mimics the first paper's very first selection). We're different in this regard.
    \item Other SMI AL retrieval work: SIMILAR (PRISM), TALISMAN, CLARIFIER.
    \item Other weak supervision work: CLARIFIER, SNORKEL, SPEAR, ...
\end{itemize}
\fi

\section{Preliminaries}
\label{sec:prelim}

First, we briefly introduce submodular functions and Submodular Mutual Information (SMI)~\cite{iyer2021submodular,iyer2021generalized}, which is used as \model's mechanism for guided active learning selection. A set function $F:2^{\Vcal} \rightarrow \mathbb{R}$ over a ground set of instances $\Vcal$ assigns a real-valued score to each possible subset of $\Vcal$. Unfortunately, finding the maximizing subset of a given size of $F$ is NP-hard. Instead, approximation algorithms for finding this subset are used. Notably, a $(1-\frac{1}{e})$-approximate greedy algorithm exists for finding the maximizing cardinality-constrained subset if $F$ is \emph{monotone submodular}~\cite{nemhauser1978analysis}. Specifically, $F$ is submodular if, for any $A\subseteq B\subseteq\Vcal$ and $a \notin B$, $F(A\cup\{a\}) - F(A) \geq F(B\cup\{a\}) - F(B)$; additionally, $F$ is monotone if $F(A\cup\{a\})-F(A)\geq 0$ for any $A$ and $a \notin A$. Such a paradigm provides a useful mechanism for selecting sets under an annotation budget within active learning by using desirable submodular functions.

\begin{table}[]
    \centering
    \caption{Instantiations of SMI functions using different submodular functions~\cite{kothawade2022prism}. Each is based on pairwise similarities $s_{ij} \in [0,1]$ between an instance $i$ and $j$.}
    \vspace{0.1in}
    \begin{tabular}{c|c}
         \hline
         \textbf{Name} & $I_F(A;Q)$ \\
         \hline \hline
         FLVMI & $\sum\limits_{i\in\Ucal} \min \left( \max\limits_{j\in A} s_{ij}, \max\limits_{j\in Q} s_{ij} \right)$ \\
         FLQMI & $\sum\limits_{i\in Q} \max\limits_{j\in A} s_{ij} + \sum\limits_{i\in A} \max\limits_{j\in Q} s_{ij}$ \\
         GCMI & $2\lambda\sum\limits_{i\in A}\sum\limits_{j\in Q} s_{ij}$ \\
         LOGDETMI & $\begin{aligned} &\log\det (S_A)\\ - &\log\det (S_A - S_{AQ}S_Q^{-1}S_{QA}) \end{aligned}$\\
         \hline
    \end{tabular}
    \label{tab:smis}
\end{table}

% https://drive.google.com/file/d/1Ccb-BGfjuhkGx5VbTk6Fn3uC2fQtipIT/view?usp=sharing
\begin{figure*}[t]
    \centering
    \includegraphics[width=\linewidth]{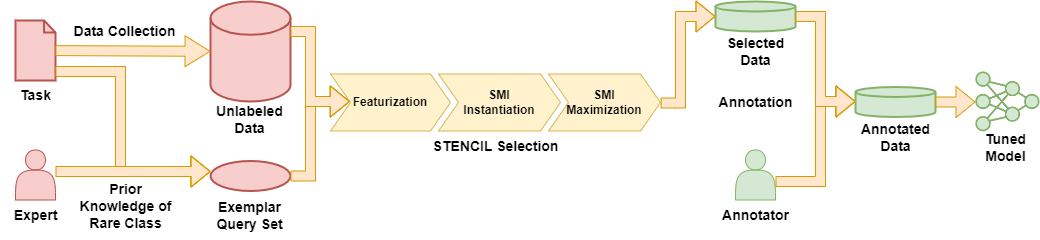}
    \caption{Architectural process of \model. A downstream task is presented, wherein subsequent data collection transpires. An expert generates an exemplar query set based on prior knowledge of the task. These artifacts are fed to \model's selection process, and the resulting selected subset is annotated and used for fine-tuning.}
    \label{fig:arch}
\end{figure*}

SMI is defined between two sets $A$ and $Q$ for a base submodular function $F$ as $I_F(A;Q) = F(A) + F(Q) - F(A\cup Q)$. Indeed,~\cite{iyer2021submodular,iyer2021generalized} note that SMI serves as a generalization of Shannon-entropic mutual information as one could recover it via SMI's definition by using Shannon entropy for $F$. Notably, if $I_F$ also satisfies certain conditions, then $I_F(A;Q)$ is also monotone submodular for a fixed $Q$~\cite{iyer2021submodular}. Accordingly, the same maximization framework discussed earlier extends to SMI.~\cite{kothawade2022prism} extend this framework for SMI by defining versions with \emph{restricted submodularity} (Table~\ref{tab:smis}); that is, $F$ is submodular for only certain subsets of $\Vcal$. By defining $\Vcal$ as $\Ucal \cup Q$ for an auxiliary set of \emph{query instances} $Q$, one can obtain a set of instances $A\subseteq\Ucal$ with high information overlap with $Q$ by maximizing $I_F(A;Q)$, providing a query-relevant $A$ that also has the salient properties (diversity, representation, etc.) of the scoring afforded by $F$. This mechanism has been used across numerous targeted active learning applications in other modalities~\cite{kothawade2021similar, kothawade2022talisman, beck2024beyond}, so its extension to text is an exciting avenue of study, especially for the cold-start scenario.

\section{Method}

In this section, we present \model, which utilizes the maximization framework mentioned in Section~\ref{sec:prelim} with a query set of exemplar instances of the rare class. We provide an overview in Figure~\ref{fig:arch} and Algorithm~\ref{alg:costal}. \model\ starts by instantiating the underlying SMI function as chosen from Table~\ref{tab:smis} through a modular choice of featurization and similarity measure. Featurization can be conducted by taking GloVe embeddings~\cite{pennington2014glove} or any similar choice. Subsequently, the similarity values required by each SMI function in Table~\ref{tab:smis} are computed using a choice of similarity measure, such as cosine similarity, RBF similarity, and so forth (coalesced as matrices of similarity values, or \emph{similarity kernels}). Once the similarity kernels are computed, a greedy monotone submodular maximization algorithm~\cite{nemhauser1978analysis, minoux2005accelerated, mirzasoleiman2015lazier,iyer2019memoization} selects a set of unlabeled instances $A\subseteq\Ucal$ of size $B$. The set $A$ is then annotated as a set $\Lcal$, and a model $M$ is then fine-tuned on $\Lcal$. In our experiments, we use an average of GloVe~\cite{pennington2014glove} embeddings and cosine similarity to instantiate each SMI function.

\begin{algorithm}
    \centering
    \caption{\model}
    \begin{algorithmic}[1]
        \INPUT Model $M$, Unlabeled set $\Ucal$, Query set $Q$, Active learning budget $B$
        \OUTPUT $M$
        \STATE $E_\Ucal, E_Q \leftarrow \textsc{Featurize}(\Ucal), \textsc{Featurize}(Q)$
        \STATE $\Kcal \leftarrow \textsc{SimilarityKernels}(E_\Ucal,E_Q)$
        \STATE $I_F \leftarrow \textsc{Instantiate}(\Kcal)$
        \STATE $A \leftarrow \emptyset$
        \WHILE{$|A| < B$}
            \STATE $a' \leftarrow \argmax\limits_{a \in \Ucal\setminus A} I_F(A\cup\{a\};Q) - I_F(A;Q)$
            \STATE $A \leftarrow A\cup\{a'\}$
        \ENDWHILE
        \STATE $\Lcal \leftarrow \textsc{Annotate}(A)$
        \STATE $M \leftarrow \textsc{FineTune}(\Lcal, M)$
    \end{algorithmic}
    \label{alg:costal}
\end{algorithm}

To apply \model, an exemplar query set is needed. The query set comes from prior knowledge of the task and contains exemplars of the rare class. While such exemplars are hard to generate for different modalities of data such as images, we posit that it is easier to derive exemplars for reasonable text-based tasks. With spam detection, for example, prior knowledge dictates that messages that urge to open hyperlinks would likely be spam instances; in such cases, one can prepare a few example sentences of this variety to use as the query set $Q$. With a complimentary choice of featurization, \model\ can then effectively choose a set of instances that semantically match the exemplars. Conducting selection in this way effectively returns a set of instances that are assumed to be \emph{weakly labeled} as the rare class, which are then \emph{strongly labeled} by an annotator.

\section{Experiments}
\label{sec:experiments}

\begin{table*}[ht]
    \centering
    \caption{Mean performance metrics with standard deviations of baselines across 10 trials. The first three rows give test accuracy (with std dev) while the last three rows give rare-class F-1 score  (with std dev).}
    \vspace{0.1in}
    \begin{tabular}{p{0.47in}|p{0.39in}p{0.39in}p{0.35in}p{0.35in}|p{0.34in}p{0.34in}p{0.34in}p{0.34in}p{0.34in}p{0.34in}p{0.4in}}
         \hline
         \textbf{Dataset} & \textbf{FLQMI} & \textbf{FLVMI} & \textbf{LDMI} & \textbf{GCMI} & \textbf{RegEx} & \textbf{Rand} & \textbf{Badge} & \textbf{Ent} & \textbf{L~Conf} & \textbf{Marg} & \textbf{KMean} \\ \hline \hline
         YouTube & 70.9 $\pm$ 6.7 & 67.4 $\pm$ 3.4 & \textbf{77.8 $\pm$ 3.2} & 58.5 $\pm$ 5.7 & 66.6 $\pm$ 4.7 & 51.5 $\pm$ 5.5 & 53.3 $\pm$ 4.1 & 52.8 $\pm$ 8 & 52.1 $\pm$ 4.8 & 51.2 $\pm$ 3.1 & 56.3 $\pm$ 2.8 \\ 
         SMS & 58.9 $\pm$ 6.9 & 66.6 $\pm$ 9.4 & 72.3 $\pm$ 4.9 & \textbf{86.9 $\pm$ 2.3} & 62.1 $\pm$ 12 & 55.8 $\pm$ 7.8 & 60.6 $\pm$ 6.9 & 68.2 $\pm$ 15 & 54.6 $\pm$ 9.4 & 60.2 $\pm$ 15 & 53.5 $\pm$ 3.6 \\ 
         Tweet & 68.4 $\pm$ 3.2 & 66.8 $\pm$ 3.1 & \textbf{72.1 $\pm$ 1.6} & 66.2 $\pm$ 3.3 & 56.2 $\pm$ 4.0 & 59.1 $\pm$ 3.0 & 56.1 $\pm$ 4.4 & 53.2 $\pm$ 4.0 & 52.4 $\pm$ 5.3 & 53.4 $\pm$ 4.1 & 62.4 $\pm$ 2.8 \\ \hline
         YouTube & 73.4 $\pm$ 5.8 & 55.1 $\pm$ 7.5 & \textbf{77.0 $\pm$ 5.9} & 61.1 $\pm$ 4.2 & 59.9 $\pm$ 7.5 & 10.0 $\pm$ 16 & 16.6 $\pm$ 13 & 12.8 $\pm$ 22 & 12.1 $\pm$ 15 & 9.7 $\pm$ 10.9 & 26.8 $\pm$ 7.6 \\ 
         SMS & 27.9 $\pm$ 20 & 46.9 $\pm$ 22 & 61.1 $\pm$ 10 & \textbf{86.9 $\pm$ 2.2} & 33.4 $\pm$ 30 & 17.5 $\pm$ 21 & 32.8 $\pm$ 19 & 46.9 $\pm$ 34 & 12.7 $\pm$ 24 & 25.3 $\pm$ 36 & 11.8 $\pm$ 11.4 \\ 
         Tweet & \textbf{72.4 $\pm$ 1.4} & 55.0 $\pm$ 7.3 & 67.2 $\pm$ 3.0 & 66.7 $\pm$ 4.4 & 24.3 $\pm$ 13 & 33.2 $\pm$ 8.9 & 24.0 $\pm$ 15 & 14.2 $\pm$ 14 & 11.2 $\pm$ 18 & 16.1 $\pm$ 14 & 43.3 $\pm$ 7.2 \\ \hline
    \end{tabular}
    \label{tab:result}
\end{table*}

\begin{table*}[t]
    \centering
    \caption{Effect of query set sizes on mean performance metrics of LOGDETMI across 10 trials. The first three rows give test accuracy (with std dev) while the last three rows give rare-class F-1 scores (with std dev). At $100\%$ query set utilization, YouTube has 15 query instances, SMS has 25 query instances, and Tweet has 20 query instances.}
    \vspace{0.1in}
    \begin{tabular}{p{0.5in}|p{0.7in}p{0.7in}p{0.7in}p{0.7in}p{0.7in}}
         \hline
         \textbf{Dataset} & \textbf{Sz-20\%} & \textbf{Sz-40\%} & \textbf{Sz-60\%} & \textbf{Sz-80\%} & \textbf{Sz-100\%} \\ \hline \hline
         YouTube & 68.8 $\pm$ 8.8 & 70.8 $\pm$ 6.0 & 74.6 $\pm$ 4.0 & 74.7 $\pm$ 2.5 & \textbf{77.2 $\pm$ 3.3} \\ 
         SMS & 63.9 $\pm$ 7.3 & 68.8 $\pm$ 4.2 & 70.9 $\pm$ 8.3 & 70.3 $\pm$ 8.4 & \textbf{73.2 $\pm$ 4.5} \\ 
         Tweet & 69.5 $\pm$ 3.7 & 70.5 $\pm$ 2.9 & 71.6 $\pm$ 1.9 & 71.9 $\pm$ 1.6 & \textbf{72.3 $\pm$ 1.6} \\ \hline
         YouTube & 58.4 $\pm$ 21.4 & 63.6 $\pm$ 12.6 & 71.9 $\pm$ 6.2 & 71.8 $\pm$ 5.9 & \textbf{76.9 $\pm$ 6.5} \\ 
         SMS & 41.5 $\pm$ 18.8 & 54.1 $\pm$ 9.5 & 57.2 $\pm$ 17.8 & 55.9 $\pm$ 20.2 & \textbf{63.1 $\pm$ 8.9} \\ 
         Tweet & 62.0 $\pm$ 8.5 & 63.6 $\pm$ 6.0 & 67.0 $\pm$ 3.7 & 67.1 $\pm$ 3.7 & \textbf{67.5 $\pm$ 2.7} \\ 
         \hline
    \end{tabular}
    \label{tab:result_abl}
\end{table*}

In this section, we experimentally verify \model's single-selection ability to handle class-imbalanced cold-start active learning scenarios for text modalities versus common active learning strategies. We additionally conduct ablations across the choice of SMI function (Table~\ref{tab:smis}) and the query set size, analyzing their effect on performance. We conduct experiments over three datasets: YouTube Spam Classification~\cite{alberto2015tubespam}, SMS Spam Classification~\cite{almeida2011contributions}, and Twitter Sentiment~\cite{wan2015ensemble} (see Appendix~\ref{app:datasets}). Such choices are natural since common knowledge of the structure of spam messages and positive sentiments can be incorporated via exemplars (shown in Appendix~\ref{app:queries}). Additionally, both of the latter datasets feature natural class imbalances (SMS: $1:18$, Twitter: $1:6$). We induce an imbalance within the YouTube dataset ($1:10$) to congregate three class-imbalanced settings, and we evaluate on balanced test data to measure both rare-class and overall performance. 

For baseline comparison with \model's SMI variants (Table~\ref{tab:smis}), we evaluate against common active learning strategies. \textbf{Random} chooses $B$ random instances from $\Ucal$. \textbf{Entropy, Least Confidence}, and \textbf{Margin} select the top $B$ instances from $\Ucal$ with highest Shannon entropy, lowest predicted probability, and lowest classification margin, respectively~\cite{settles2009active}. \textbf{Badge} calculates the loss gradients of each point in $\Ucal$ using pseudo-labeling and performs \textsc{k-means++} sampling~\cite{arthur2007k} to select $B$ instances from $\Ucal$, which gives a diverse set of uncertain instances~\cite{ash2019deep}. \textbf{RegEx} uses regular expression matching between each data point in the exemplar query set and each data point $\Ucal$. The top $B$ samples with the highest cumulative count of word matches with the patterns or phrases in the query set are selected. Finally, \textbf{KMean} applies $k$-means clustering with $k=B$ on the space of average GloVe embeddings for each instance and returns the 1-NN to each center, which serves as an adjacent baseline to~\cite{hacohen2022active, jin2022cold}. In all cases, $B$ is set to 50 for YouTube, 136 for SMS, and 144 for Twitter (roughly $1.5\%$ to $5\%$ of the full unlabeled set size for each dataset). Subsequently, we utilize an LSTM-based network architecture~\cite{hochreiter1997long} of $1.23M$ parameters to take the sequence as input and make classification decisions. We optimize the network utilizing SGD with learning rate $0.01$ for 25-50 epochs (YouTube: 50, SMS: 30, Tweet: 25).

We present the results of our experiments averaged across 10 trials in Table~\ref{tab:result} and Table~\ref{tab:result_abl}. Summarily, we see that the GCMI variant of \model\ performs the best on the SMS dataset, the LOGDETMI variant performs the best on the YouTube dataset, and the LOGDETMI variant exhibits the best overall accuracy (on balanced test data) while the FLQMI variant exhibits the best rare-class F-1 score on the Twitter dataset. \emph{Hence, \model\ improves the rare-class performance in the class-imbalanced cold-start setting using only one round of selection without sacrificing overall performance}, seeing improvements of $10\%-18\%$ overall accuracy and $17\%-40\%$ rare-class F-1 score. Interestingly, we note that while RegEx uses the exemplar query set, its rule-based operation cannot capture diverse sets of semantically related instances as well as \model. Indeed, LOGDETMI accounts for diversity through its determinants, which is advantageous in cold-start settings where a good coverage of the data is required. GCMI's effectiveness is particularly notable in situations where rare instances exhibit high similarity, as observed in the SMS dataset. Lastly, we observe from Table~\ref{tab:result_abl} that, while more query examples provides better performance, there is a diminishing return on investment in terms of performance gains when increasing the query set size beyond a certain point, which means that \emph{\model\ effectively leverages small query sets and does not specifically require large query sets}.

\section{Conclusion}

In this work, we present \model, which effectively utilizes SMI and prior knowledge of the data environment in the form of a small exemplar set of rare-class instances to manage class-imbalanced cold-start active learning scenarios. We demonstrate that \model\ is able to make improvements of $10\%-18\%$ overall accuracy and $17\%-40\%$ rare-class F-1 score over common active learning strategies within a single round of selection. As an exciting avenue of future work, the application of \model's SMI functionality to other text-based tasks beyond classification -- and the formation of query exemplars in these tasks -- would greatly expand the contributions of this work.\looseness-1

\section*{Reproducibility and Licenses}

To reproduce our results, we offer six Google Colab notebooks\footnote{\url{https://github.com/nab170130/stencil}} that can be run repeatedly to amass trial outcomes for each configuration of \model\ and for each baseline method. Per the setting and model size ($1.23M$ parameters) described in Section~\ref{sec:experiments}, it is sufficient to run these notebooks on the standard CPU configuration. Per dataset, amassing 10 trials requires roughly 2 hours of compute time. Hyper-parameters were chosen based on the code examples provided in DISTIL.

We utilize the following libraries in our experiments and list their licenses as follows:

\begin{itemize}
    \item DISTIL~\cite{beck2021effective}: MIT License
    \item Submodlib~\cite{kaushal2022submodlib}: MIT License
    \item PyTorch~\cite{paszke2019pytorch}: Modified BSD License
    \item NLTK~\cite{bird2009natural}: Apache License Version 2.0
    \item Transformers~\cite{wolf2020transformers}: Apache License 2.0
    \item YouTube Spam~\cite{alberto2015tubespam}: CC Attribution 4.0 International
    \item SMS Spam~\cite{almeida2011contributions}: CC Attribution 4.0 International
    \item Twitter Sentiment~\cite{wan2015ensemble}: Creative Commons CC-BY-NC-ND
\end{itemize}

\bibliography{DMLR@ICML24/main}
\bibliographystyle{icml2024}

%%%%%%%%%%%%%%%%%%%%%%%%%%%%%%%%%%%%%%%%%%%%%%%%%%%%%%%%%%%%%%%%%%%%%%%%%%%%%%%
%%%%%%%%%%%%%%%%%%%%%%%%%%%%%%%%%%%%%%%%%%%%%%%%%%%%%%%%%%%%%%%%%%%%%%%%%%%%%%%
% APPENDIX
%%%%%%%%%%%%%%%%%%%%%%%%%%%%%%%%%%%%%%%%%%%%%%%%%%%%%%%%%%%%%%%%%%%%%%%%%%%%%%%
%%%%%%%%%%%%%%%%%%%%%%%%%%%%%%%%%%%%%%%%%%%%%%%%%%%%%%%%%%%%%%%%%%%%%%%%%%%%%%%
\newpage
\appendix
\onecolumn

\section{Chosen Query Exemplars within Experiments}
\label{app:queries}

In this section, we provide the query exemplars used in Section~\ref{sec:experiments}. The query phrases are created using domain knowledge, examples from prior work~\cite{maheshwari-etal-2022-learning}, and prompting ChatGPT~\cite{blog2023introducing} with the following: "Give me 10 example phrases, not sentences, found in each of 1. Spam youtube comments 2. Spam SMS messages 3. Tweets with positive sentiment." Notably, we did not examine the instances of YouTube Spam~\cite{alberto2015tubespam}, SMS Spam~\cite{almeida2011contributions}, or Twitter Sentiment~\cite{wan2015ensemble} when curating these query sets. 

\subsection{YouTube Spam}
\begin{itemize}
    \item `check out my latest video' 
    \item `click the link'
    \item `dont miss out'
    \item `https'
    \item `for more information'
    \item `free gift'
    \item `free giveaway'
    \item `win prizes'
    \item `like and comment'
    \item `limited time offer'
    \item `please help'
    \item `subscribe now'
    \item `subscribe to my channel'
    \item `visit my website'
    \item `watch my video'
\end{itemize}

\subsection{SMS Spam}
\begin{itemize}
    \item "Click this link to claim your cash prize"
    \item "Confirm your account details to continue using our service"
    \item "Congratulations! You've won a free trial"
    \item "Earn money quickly"
    \item "Exclusive discount code inside"
    \item "Free gift awaiting you" 
    \item "Get paid to work from home"
    \item "Important security notice"
    \item "Problem with your payment method"
    \item "Special limited-time offer" 
    \item"Suspicious activity detected"
    \item "To stop receiving these messages, click here"
    \item "Unclaimed money in your name"
    \item "Urgent action required to receive your package"
    \item "Verify your email and password immediately"
    \item "Warning: Your account will be deactivated"
    \item "You are eligible for a refund" 
    \item "You have an unpaid bill"
    \item "You're our lucky shopper today"
    \item "You've been selected for an exclusive offer"
    \item "Your account has been temporarily locked"
    \item "Your order is ready for pickup"
    \item "Your subscription is about to expire"
    \item "Your trial period is ending"
    \item "Your warranty is expired"
\end{itemize}

\subsection{Twitter Sentiment}
\begin{itemize}
    \item "Absolutely love"
    \item "Amazing job"
    \item "Can't wait for"
    \item "Extremely happy to"
    \item "Feeling blessed"
    \item "Feeling inspired by"
    \item "Feeling optimistic about"
    \item "Feeling very proud of"
    \item "Had a great time"
    \item "Highly recommend"
    \item "Incredible experience"
    \item "Overwhelmed with happiness"
    \item "Really excited about"
    \item "So grateful for"
    \item "Such a beautiful"
    \item "Thank you so much"
    \item "Totally loving the"
    \item "Truly amazing"
    \item "Very successful"
    \item "Wonderful day"
\end{itemize}

\section{Dataset Details}
\label{app:datasets}

In this section, we provide additional details of each dataset used in Section~\ref{sec:experiments}. All datasets contain English text. We also provide class distributions in our train and test splits in Table~\ref{tab:splits}.

\begin{table*}[]
    \centering
    \caption{Class distributions used in the train-test splits of our experiments (Section~\ref{sec:experiments}). The train split is made unlabeled before selection occurs.}
    \begin{tabular}{l|cc|cc}
         \hline
         \textbf{Dataset} & \textbf{Rare Train} & \textbf{Common Train} & \textbf{Rare Test} & \textbf{Common Test} \\
         \hline \hline 
         YouTube & 85 & 808 & 151 & 143\\
         SMS & 234 & 4312 & 480 & 476\\
         Tweet & 1402 & 8178 & 936 & 909\\
         \hline
    \end{tabular}
    \label{tab:splits}
\end{table*}

\textbf{YouTube Spam Classification}~\cite{alberto2015tubespam}: A collection of real comments on five of the ten most viewed YouTube videos at the time of collection. It consists of 1005 comments marked spam and 951 comments marked ham (non-spam). This dataset is publicly available in the UCI Machine Learning Repository\footnote{\url{https://archive.ics.uci.edu/dataset/380/youtube+spam+collection}}. After creating a balanced test set and making the spam class as the rare class in the training data, we induce a class imbalance of approximately $1:10$.

\textbf{SMS Spam Classification}~\cite{almeida2011contributions}: A dataset of SMS messages curated for binary spam classification task. It combines 425 SMS spam messages from the Grumbletext website, a subset of 3,375 ham messages from the NUS SMS Corpus, 450 ham messages from a PhD Thesis, and 1,002 ham and 322 spam messages from the SMS Spam Corpus v.0.1 Big. This dataset is publicly accessible at the UCI ML Repository\footnote{\url{https://archive.ics.uci.edu/dataset/228/sms+spam+collection}}. After creating a balanced test set, the imbalance factor is left as approximately $1:18$.

\textbf{Twitter Sentiment}~\cite{wan2015ensemble}: A sentiment classification dataset of comments from Twitter (tweets) about airline services. It includes 12,864 tweets labeled with positive, neutral, or negative sentiments based on analysis using an ensemble of 6 classifiers. For the scope of this study, only entries with positive and negative labels have been retained. The dataset is made available in~\cite{maheshwari-etal-2022-learning}. After creating a balanced test set, the imbalance factor is left as approximately $1:6$ (with positive sentiments deemed rare).

\section{Additional Baseline Details}
\label{app:baselines}

In this section, we provide additional detail for each baseline used in our experiments.

\subsection{Submodular Mutual Information}

As mentioned in Section~\ref{sec:prelim}, SMI can be instantiated utilizing various submodular functions, each of which models different properties. Facility Location ($F(A) = \sum_{i\in\Vcal} \max_{j\in A} s_{ij}$) captures representation information between a subset and its ground set and is used to instantiate the FLVMI and FLQMI variants in Table~\ref{tab:smis}. As derived in~\cite{kothawade2022prism}, FLQMI differs from FLVMI by modeling only the cross similarities between $A$ and $Q$. Log Determinant ($F(A) = \log\det S_A$) captures diversity information in $A$, which is reflected by the determinant of $S_A$; hence, LOGDETMI captures diversity information between $A$ and $Q$. Graph Cut ($F(A) = \sum_{i\in\Vcal}\sum_{j\in A} s_{ij} - \lambda \sum_{i,j\in A} s_{ij}$) also focuses on representation of $\Vcal$ with $A$ and, when ingested by SMI's definition, gives GCMI, which focuses entirely on how relevant $A$ is to the queries in $Q$. Notably, all instantiations utilize similarity values between instances. In our experiments, we represent each instance via the average of the GloVe~\cite{pennington2014glove} embeddings for each token in the sequence. Afterwards, we compute cosine similarity to derive each $s_{ij}$ value.

\subsection{BADGE Sampling}

The BADGE strategy~\cite{ash2019deep} is a common choice for batch active learning where diverse sets of uncertain instances are selected, which tend to be informative instances. BADGE achieves this by embedding each unlabeled instance using hypothesized loss gradients (using the most confidently predicted class). By picking a diverse span of points with large magnitude gradients, one obtains a diverse set of instances that are likely to bring large updates to the loss (and are thus uncertain). In general, the following steps are taken:
\begin{itemize}
    \item Calculate the pseudo-label for each point in the unlabeled set. The pseudo-label is the class with the highest probability.
    \item Compute the cross-entropy loss for each point in the unlabeled set using this pseudo-label.
    \item Obtain the resulting loss gradients on the last linear layer of the model for each point (the hypothesized loss gradients).
    \item Using these gradients as a form of embedding for each unlabeled point, run \textsc{k-means++} initialization~\cite{arthur2007k} on this embedding set, retrieving $B$ centers. Each center is a point from the unlabeled set, and $B$ represents the active learning budget.
    \item Request labels for the $B$ points whose embeddings were selected.
\end{itemize}

\subsection{Basic Uncertainty Sampling}

A common and long-standing choice for active learning selection is that of simple uncertainty sampling measures as discussed in~\cite{settles2009active}. Namely, one can quantify uncertainty utilizing the predicted class probabilities and select the top $k$ most uncertain points. Typically, there are three common choices for quantifying uncertainty: \textbf{Entropy} ($H(x) = -\sum_i p(x)_i\log p(x)_i$), \textbf{Least Confidence} ($C(x) = \max_i p(x)_i$), and \textbf{Margin} ($M(x) = p(x)_{\sigma_1} - p(x)_{\sigma_2}$, where $\sigma_1$ and $\sigma_2$ denote the most probable and second-most probable class, respectively). \textbf{Entropy sampling} chooses the top $B$ $H(x)$ values. \textbf{Least confidence sampling} and \textbf{margin sampling} choose the $B$ smallest $C(x)$ and $M(x)$ values, respectively. The uncertain points are then labeled, which tends to provide more crucial information to the downstream model.

\subsection{RegEx Sampling}

Given the list of query phrases specific to the dataset, the RegEx Selection method implements a regular expression matching between each data point in the query set and each data point in the unlabeled dataset. The samples with the highest cumulative count of matches with the patterns or phrases in the query set make the selected subset of the training dataset. Note that this sampling strategy focuses only on extracting rare class samples and not on obtaining a balanced representation of both class labels in the dataset.

\section{Limitations and Societal Impacts}

Here, we briefly discuss the limitations of \model. Namely, proper function of \model\ is contingent upon the \emph{correct} prior knowledge being injected as the query set. If improper prior knowledge is used, \model\ may not select the rare-class data optimally. As such, \model's application is limited to settings where adequate query exemplars can be created \emph{a priori}, which is not the case for many settings. As a text example, tasks that involve processing SMILES-string representations of chemicals may not be easy to provide exemplars. Further, other modalities may not be as easy to provide exemplars as text modalities, such as video, images, audio, and so forth. Hence, \model\ is limited to tasks where query exemplars are easily provided. Another limitation of \model\ is that its core functionality depends on an adequate feature space within the cold-start setting. While this is typically the case with NLP tasks via the advent of pre-trained models, certain tasks may not be compatible with the embedding space for these models. Hence, \model\ additionally needs access to good featurization of the data within a cold-start setting.

While we do not anticipate this work to have far-reaching societal consequences, we do highlight that this paper provides a query-biased basis for active learning methods within the cold-start setting. Hence, as \model\ provides a base dataset upon which other active learning methods can build, it is important to mitigate harmful biases that might be present within the query data. Otherwise, these biases may be propagated in downstream learning, which can result in biased models that inadvertently mishandle important domains of the input space. We believe this to be the main potential consequence of this work and urge future applications to take precautions to mitigate this effect.

\end{document}